\documentclass[journal]{IEEEtran}
\usepackage{amsfonts}
\usepackage{mathrsfs}
\usepackage{cite}
\usepackage{graphicx}
\usepackage{amsmath}
\usepackage{booktabs}   
\usepackage[linesnumbered,ruled,vlined]{algorithm2e}
\usepackage[usenames, dvipsnames]{color}
\usepackage{soul}
\usepackage{bm}

\begin{document}

\title{Multi-Modal Multi-Scale Deep Learning for Large-Scale Image Annotation}

\author{Yulei~Niu,~Zhiwu~Lu,~Ji-Rong~Wen,~Tao Xiang,~and~Shih-Fu~Chang%
\thanks{Y. Niu, Z. Lu, and J.-R. Wen are with the Beijing Key Laboratory of Big Data Management and Analysis Methods, School of Information, Renmin University of China, Beijing 100872, China (email: niu@ruc.edu.cn,~luzhiwu@ruc.edu.cn,~jrwen@ruc.edu.cn).}
\thanks{T. Xiang is with the School of Electronic Engineering and Computer Science, Queen Mary University of London, Mile End Road, London E1 4NS, United Kingdom (email: t.xiang@qmul.ac.uk).}
\thanks{S.-F. Chang is with the Department of Electrical Engineering, Columbia University, New York, NY 10027, USA. (email: sc250@columbia.edu).}
}

\maketitle

\begin{abstract}
Image annotation aims to annotate a given image with a variable number of class labels corresponding to diverse visual concepts. In this paper, we address two main issues in large-scale image annotation: 1) how to learn a rich feature representation suitable for predicting a diverse set of visual concepts ranging from object, scene to abstract concept; 2) how to annotate an image with the optimal number of class labels. To address the first issue, we propose a novel multi-scale deep model for extracting rich and discriminative features capable of representing a wide range of visual concepts. Specifically, a novel two-branch deep neural network architecture is proposed which comprises a very deep main network branch and a companion feature fusion network branch designed for fusing the multi-scale features computed from the main branch. The deep model is also made multi-modal by taking noisy user-provided tags as model input to complement the image input. For tackling the second issue, we introduce a label quantity prediction auxiliary task to the main label prediction task to explicitly estimate the optimal label number for a given image. Extensive experiments are carried out on two large-scale image annotation benchmark datasets and the results show that our method significantly outperforms the  state-of-the-art.
\end{abstract}

\begin{IEEEkeywords}
Large-scale image annotation, multi-scale deep model, multi-modal deep model, label quantity prediction.
\end{IEEEkeywords}

%
\IEEEpeerreviewmaketitle

\section{Introduction}

\IEEEPARstart{I}mage recognition \cite{Khotanzad2002Invariant,Boureau2011Ask,Lu2010Image,Lu2009Image} is a fundamental problem in image content analysis. It aims to assign a single class label to a given image which typically corresponds to a specific type of visual concept, e.g., the (single) object contained in the image or the visual scene depicted by the image. In the past five years, significant advances have been achieved on large-scale image recognition tasks \cite{he2016cvpr,krizhevsky2012nips,simonyan2014arxiv,szegedy2015cvpr,SI17}, with the latest models being capable of beating humans on these visual recognition tasks. These advances are mainly attributed to the existence of large-scale benchmarks such as ImageNet \cite{Russakovsky2014ImageNet} and the deployment of a deep representation learning paradigm based on  deep neural networks (DNNs) which learn the optimal representation and classifier jointly in an end-to-end fashion.

As a closely related task, image annotation \cite{Makadia2008A,Russell2008LabelMe,Lu2011Contextual,Lu2015Semantic,Jing2016Multi} aims to describe (rather than merely recognise) an image by annotating all visual concepts that appear in the image. This brings about a number of differences and new challenges compared to the image recognition problem. First, image annotation is a multi-label multi-class classification problem \cite{Tsoumakas2007Multi,Read2011Classifier}, instead of a single-label multi-class classification problem as in image recognition. The model thus needs to predict multiple labels for a given images with richer contents than those used in image recognition. The second difference, which is more subtle yet more significant, is on the label types: labels in image annotation can refer to a much wider and more diverse range of visual concepts including scene properties, objects, attributes, actions, aesthetics etc. This is because image annotation benchmark datasets have become larger and larger \cite{Weston2010Large,Chen2010Efficient,Tsai2012Large,Feng2014Large} by crawling images and associated noisy user-provided tags from social media websites such as Flickr. These images are uploaded by a large variety of users capturing an extremely diverse set of visual scenes containing a wide range of visual concepts. Finally, in image annotation, the task is to predict not only more than one label, but also a variable number of labels -- an image with simple content may require only one or two labels to describe, whilst more complicated content necessitates more labels. In summary, these three differences make image annotation a much harder problem far from being solved.

\begin{figure}
\centering
\includegraphics[width=0.96\columnwidth]{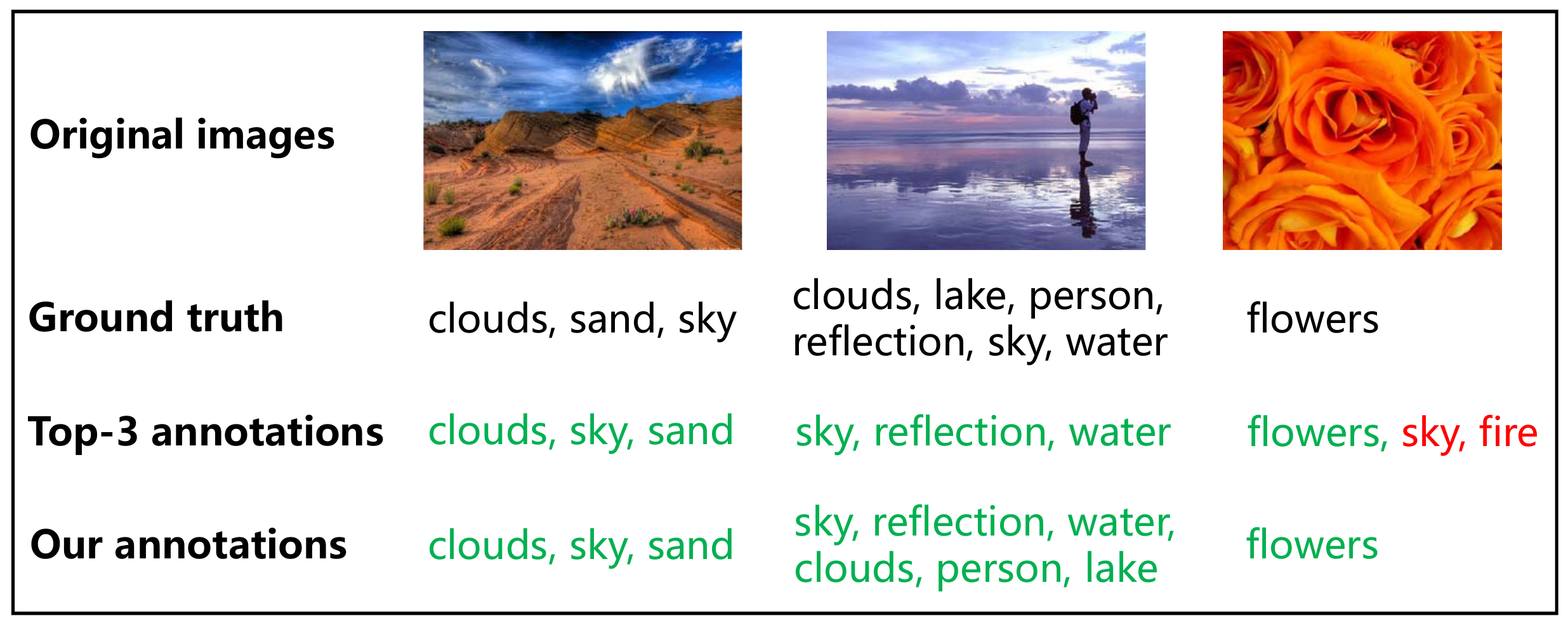}
\vspace{-0.1in}
\caption{Illustration of the difference between the traditional image annotation models that generate top-\emph{k} annotations (e.g. $k=3$) and our model that generates an automatically-determined number of annotations. Annotations with green font are correct, while annotations with red font are wrong.}\label{fig1}
\vspace{-0.1in}
\end{figure}

Existing methods on image annotation focus on solving the multi-label classification problem. Two lines of research exist. In the first line, the focus is on modeling the correlation of different labels for a given image (e.g. cow typically co-exists with grass) \cite{hu2016cvpr} to make the multi-label prediction more robust and accurate. In the second line, side information such as the noisy user-provided tags are used as additional model input from a different modality (i.e., text) \cite{johnson2015iccv}. These models ignore the variable label number problem and simply predict the top-\textit{k} most probable labels per image \cite{Makadia2008A,Russell2008LabelMe,Lu2011Contextual,
Lu2015Semantic,Jing2016Multi}. This is clearly sub-optimal as illustrated in Fig.~\ref{fig1}.

\begin{figure*}[t]
\centering
\includegraphics[width=0.99\textwidth]{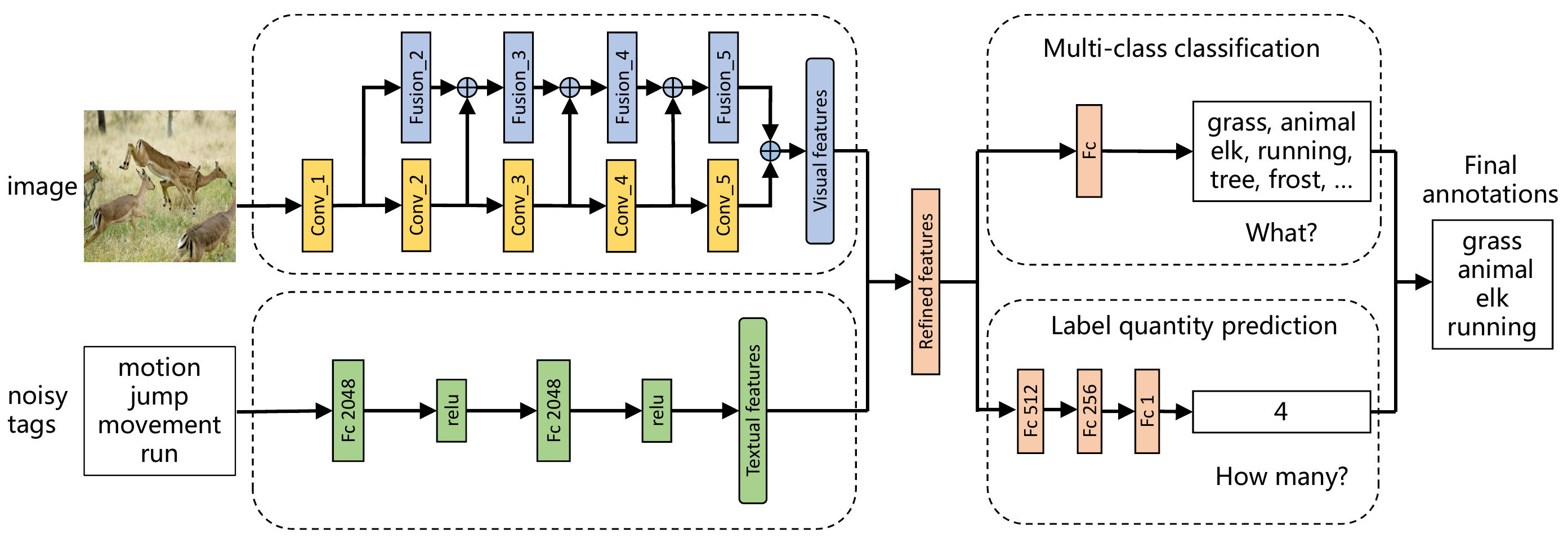}
\vspace{-0.00in}
\caption{The flowchart of our multi-modal multi-scale deep learning model for large-scale image annotation. In this model, four components are included: visual feature learning, textual feature learning, multi-class classification, and label quantity prediction. }\label{fig2}
\vspace{-0.05in}
\end{figure*}

More recently the variable label number problem has been identified and a number of solutions have been proposed \cite{liu2016arxiv,jin2016icpr,wang2016cvpr}. These solutions treat the image annotation problem as an image to text translation problem and solve it using an encoder-decoder model. Specifically, the image content is encoded using a convolutional neural network (CNN) and then fed into a recurrent neural network (RNN) \cite{Graves2009A,Li2015Constructing,Gregor2015DRAW} which outputs a label sequence. The RNN model is able to automatically determine the length of the label sequence thus producing variable numbers of labels for different images. However, one fundamental limitation of these existing approaches is that the original training labels are orderless whilst the RNN model requires a certain output label order for training. Existing methods have to introduce artificial orders such as rarest or more frequent label first. This indirect approach thus leads to sub-optimal estimation of the label quantity.

Nevertheless the most important difference and challenge, i.e. the rich and diverse label space problem, has never been tackled explicitly. In particular, recent image annotation models use deep learning in various ways: they either directly apply a CNN pretrained on the ImageNet image recognition task to extract features followed by a non-deep multi-label classification model \cite{johnson2015iccv}; or fine-tune a pretrained CNN on the image annotation benchmark datasets and obtain both the feature representation and classifier jointly \cite{gong2013arxiv,hu2016cvpr}. However, all the CNN models used were  designed originally for the image recognition task. Specifically, only the final layer feature output is used as input to the classifier. It has been demonstrated \cite{gong2014eccv,yang2015iccv,cai2016eccv,LZ16} that features computed by  a deep CNN correspond to visual concepts of higher levels of abstraction  when progressing from bottom to top layer feature maps, e.g., filters learned in bottom layers could represent colour and texture whilst those in top layers object parts. This means that only the most abstract features were used for the classifier in the existing models. However, as mentioned earlier, for image annotation, the visual concepts to be annotated/labeled have drastically different levels of abstraction; for instance, `grass' and `sand' can be sufficiently described by colour and texture oriented features computed at the bottom layers of a CNN, whist  `person', `flower' and `reflection' are more abstract thus requiring features learned from top layers.  These image recognition-oriented CNNs, with or without fine-tuning, are thus unsuitable for the image annotation task because they fail to provide rich representations at different abstraction scales.

In this paper, we propose a novel multi-modal multi-scale deep neural network (see Fig.~\ref{fig2}) to address both the diverse label space problem and variable label number problem. First, to extract and fuse features learned at different scales, a novel deep CNN architecture is proposed. It comprises two network branches: a main CNN branch which can be any contemporary CNN such as GoogleNet \cite{szegedy2015cvpr} or ResNet \cite{he2016cvpr}; and a companion multi-scale feature fusion branch which fuses features extracted at different layers of the main branch whilst maintaining the feature dimension as well as performing automated feature selection. Second, to estimate the optimal number of labels that is needed to annotate a given image, we explicitly formulate an optimum label quantity estimation task and optimize it jointly with the label prediction main task. Finally, to further improve the label prediction accuracy, the proposed model is made multi-modal by taking both the image and noisy user-provided tags as input.

Our main contributions are as follows: (1) A novel multi-scale deep CNN architecture is proposed which is capable of effectively extracting and fusing features at different scales corresponding to visual concepts of different levels of abstraction. (2) The variable label number problem is solved by explicitly estimating the optimum label quantity in a given image. (3) We also formulate a multi-modal extension of the model to utilize the noisy user-provided tags. Extensive experiments are carried out on two large-scale benchmark datasets: NUS-WIDE \cite{chua2009civr} and MSCOCO \cite{vinyals2017pami}. The experimental results demonstrate that our model outperforms  the state-of-the-art alternatives, often by a significant margin.

\section{Related Work}

\subsection{Multi-Scale CNN Models}

Although multi-scale representation learning has never been attempted for image annotations, there are existing efforts on designing CNN architectures that enable multi-scale feature fusion. Gong \textit{et al.} \cite{gong2014eccv} noticed that the robustness of global features was limited due to the lack of geometric invariance, and thus proposed a multi-scale orderless pooling (MOP-CNN) scheme, which concentrates orderless Vectors of Locally Aggregated Descriptors (VLAD) pooling of CNN activations at multiple levels. Yang and Ramanan \cite{yang2015iccv} argued that different scales of features can be used in different image classification tasks through multi-task learning, and developed directed acyclic graph structured CNNs (DAG-CNNs) to extract multi-scale features for both coarse and fine-grained classification tasks. Cai \textit{et al.} \cite{cai2016eccv} presented a multi-scale CNN (MS-CNN) model for fast object detection, which performs object detection using both lower and higher output layers. Liu \textit{et al.} \cite{LZ16} proposed a multi-scale triplet CNN model for person re-identification. The results reported in \cite{gong2014eccv,yang2015iccv,cai2016eccv,LZ16} have shown that multi-scale features are indeed effective for image content analysis. In this paper, a completely new network architecture is designed by introducing the companion feature fusion branch. One of the key advantage of our architecture is that the feature fusion takes place in each layer without increasing the feature dimension, thus maintaining the same final feature dimension as the main branch. This means that our architecture can adopt arbitrarily deep network in the branch without worrying about the explosion of the fused feature dimension.

\subsection{Label Quantity Prediction for Image Annotation}

Early works \cite{hu2016cvpr,johnson2015iccv,gong2013arxiv} on image annotation  assign top-$k$ predicted class labels to each image. However, the quantities of class labels in different images vary significantly, and the top-$k$ annotations degrade the performance of image annotation. To overcome this issue, recent works start to predict variable numbers of class labels for different images. Most of them adopt a CNN-RNN encoder-decoder architecture, where the CNN subnetwork encodes the input pixels of images into visual feature, and the RNN subnetwork decodes the visual feature into a label prediction path \cite{jin2016icpr,liu2016arxiv,wang2016cvpr}. Specifically, RNN can not only perform classification, but also control the number of output class labels -- the model stop emitting labels once an end-of-label-sequence token is predicted. Since RNN requires an ordered sequential list as input, the unordered label set is expected to be transformed in advance based on heuristic rules such as the rare-first rule \cite{wang2016cvpr} or the frequent-first rule \cite{jin2016icpr}. These artificially imposed rules can lead to suboptimal estimation of the label quantity. In contrast, in this work our model directly estimate the label quantity without making any assumption on the label order.

\subsection{Image Annotation Using Side Information}

A number of recent works attempted to improve the label prediction accuracy by exploiting side information collected from social media websites where the benchmark images were collected. These user-provided side information can be extracted from the noisy tags \cite{johnson2015iccv,liu2016arxiv,hu2016cvpr} and group labels \cite{hu2016cvpr,Ulges2011Learning,Zhu2009Automatic}. Johnson \textit{et al.} \cite{johnson2015iccv} found neighborhoods of images nonparametrically according to the image metadata, and combined the visual features of each image and its neighborhoods. Liu \textit{et al.} \cite{liu2016arxiv} filtered noisy tags to obtain true tags, which serve as the supervision to train the CNN model and also set the initial state of RNN. Hu \textit{et al.} \cite{hu2016cvpr} utilized both tags and group labels to form a multi-layer group-concept-tag graph, which can encode the diverse label relations. Different from \cite{hu2016cvpr} that adopted a deep model, the group labels were used to learn the context information for image annotation in \cite{Ulges2011Learning,Zhu2009Automatic}, but without considering deep models. In this paper, we introduce an additional network branch which takes noisy tags as input and the extracted textual features are fused with the visual features extracted by CNN models to boost the annotation accuracy. Moreover, since user tagged labels may be incorrect or incomplete, some recent works have focused on training with noisy labels. Liu \textit{et al.} \cite{liu2016classification} use importance reweighting to modify any loss function for classification with noisy labels, and extend the random classification noise to a bounded case\cite{cheng2017learning}. Moreover, Yu \textit{et al.} \cite{yu2017learning} prove that biased complementary labels can enhance multi-class classification. These methods have been successfully applied in binary classification and multi-class classification, which may lead to improvements when extended to our large-scale image annotation where each image is annotated with one or more tags and labels (but out of the scope of this paper).

\section{The Proposed Model}

As shown in Fig.~\ref{fig2}, Our multi-modal multi-scale deep model for large-scale image annotation consists of four components: visual feature learning, textual feature learning, multi-class classification, and label quantity prediction. Specifically, a multi-scale CNN subnetwork is proposed to extract visual feature from raw image pixels, and a multi-layer perception subnetwork is applied to extract textual features from noisy user-provided tags. The joint visual and textual features are connected to a simple fully connected layer for multi-class classification. To determine the optimum number of labels required to annotate a given image, we utilize another multi-layer perception subnetwork to predict the quantity of class labels. The results of multi-class classification and label quantity prediction are finally merged for image annotation. In the following, we will first give the details of the four components of our model, and then provide the algorithms for model training and test process.

\begin{figure}
\centering
\includegraphics[width=0.79\columnwidth]{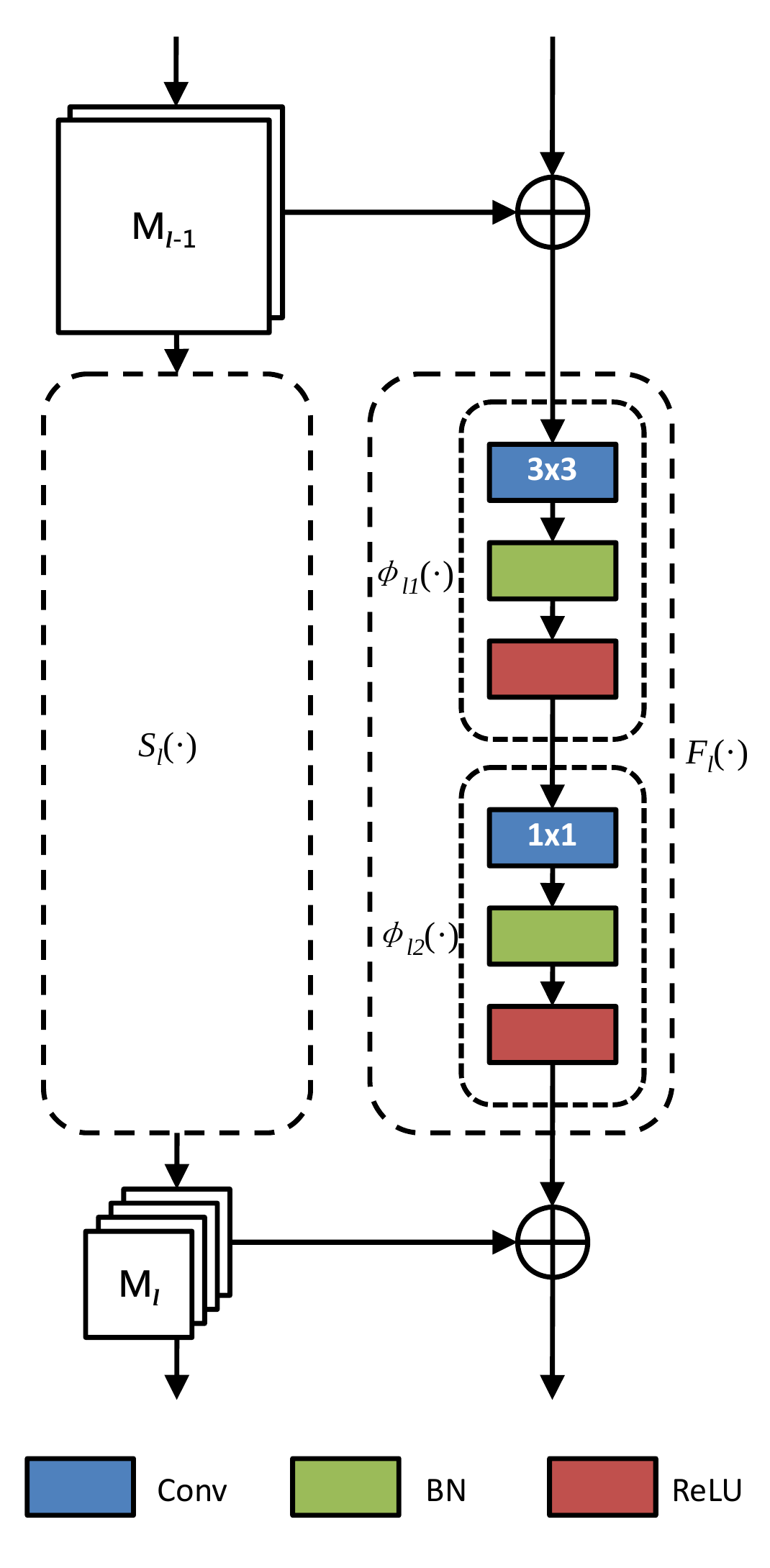}
 \caption{The block architecture of the feature fusion model.}\label{fig3}
\end{figure}

\subsection{Multi-Scale CNN for Visual Feature Learning}

Our multi-scale CNN subnetwork consists of two branch (see Fig.~\ref{fig2}): the main CNN branch which encodes an image to both low-level and high-level features, and the feature fusion branch which fuses multi-scale features extracted from the main branch.\\
\textbf{Main CNN Branch.} Given the raw pixels of an image $\textbf{I}$, the basic CNN branch encodes them into $K$ levels of feature maps $\bm{M}_1, \bm{M}_2, \cdots,\bm{M}_K$ through a series of scales $S_1(\cdot), S_2(\cdot), \cdots, S_K(\cdot)$. Each scale can be a composite function of operations, such as convolution (Conv), pooling (Pooling), batch normalization (BN), and activation function. The encoding process can be formulated as:
\begin{equation}\label{eq1}
\begin{aligned}
        &\bm{M}_0=\textbf{I}\\
        &\bm{M}_l=S_l(\bm{M}_{l-1}), l=1,2,...,K
\end{aligned}
\end{equation}
For the main CNN branch, the last feature map $\bm{M}_K$ is often used to produce the final feature vector, e.g., the conv5\_3 layer of the ResNet-101 model \cite{he2016cvpr}.\\
\textbf{Feature Fusion Branch.} When multi-scale feature maps $\bm{M}_1, \bm{M}_2, \cdots, \bm{M}_K$ are obtained by the main branch, the feature fusion model is proposed to combine these original feature maps iteratively via a set of fusion functions $\left\{F_l\right(\cdot)\}$, as shown in Fig.~\ref{fig3}. The fused feature map $\widetilde{\bm{M}}_l$ is formulated as follows:
\begin{equation}\label{eq2}
\begin{aligned}
&\widetilde{\bm{M}}_1=\bm{M}_1\\
&\widetilde{\bm{M}}_l=\bm{M}_l+F_l(\widetilde{\bm{M}}_{l-1}), l=2,...,K
\end{aligned}
\end{equation}
In this paper, we define the fusion function $F_{l}(\cdot)$ as:
\begin{equation}\label{eq3}
F_{l}(\cdot)=\phi_{l2}\left(\phi_{l1}\left(\cdot\right)\right)
\end{equation}
where $\phi_{l1}(\cdot)$ and $\phi_{l2}(\cdot)$ are two composite functions consisting of three consecutive operations: a convolution (Conv), followed by a batch normalization (BN) and a rectified linear unit (ReLU). The difference between $\phi_{l1}(\cdot)$ and $\phi_{l2}(\cdot)$ lies in the convolution layer. The $3\times3$ Conv in $\phi_{l1}(\cdot)$ is used to guarantee that $\bm{M}_l$ and $F_l(\widetilde{\bm{M}}_{l-1})$ have the same height and weight, while the $1\times1$ Conv in $\phi_{l2}(\cdot)$ can not only increase dimensions and interact information between different channels, but also reduce the number of parameters and improve computational efficiency \cite{szegedy2015cvpr,he2016cvpr}. At the end of the fused feature map $\widetilde{\bm{M}}_K$, an average pooling layer is used to extract the final visual feature vector $\bm{f}_v$ for image annotation.

\begin{figure}
\centering
\includegraphics[width=0.995\columnwidth]{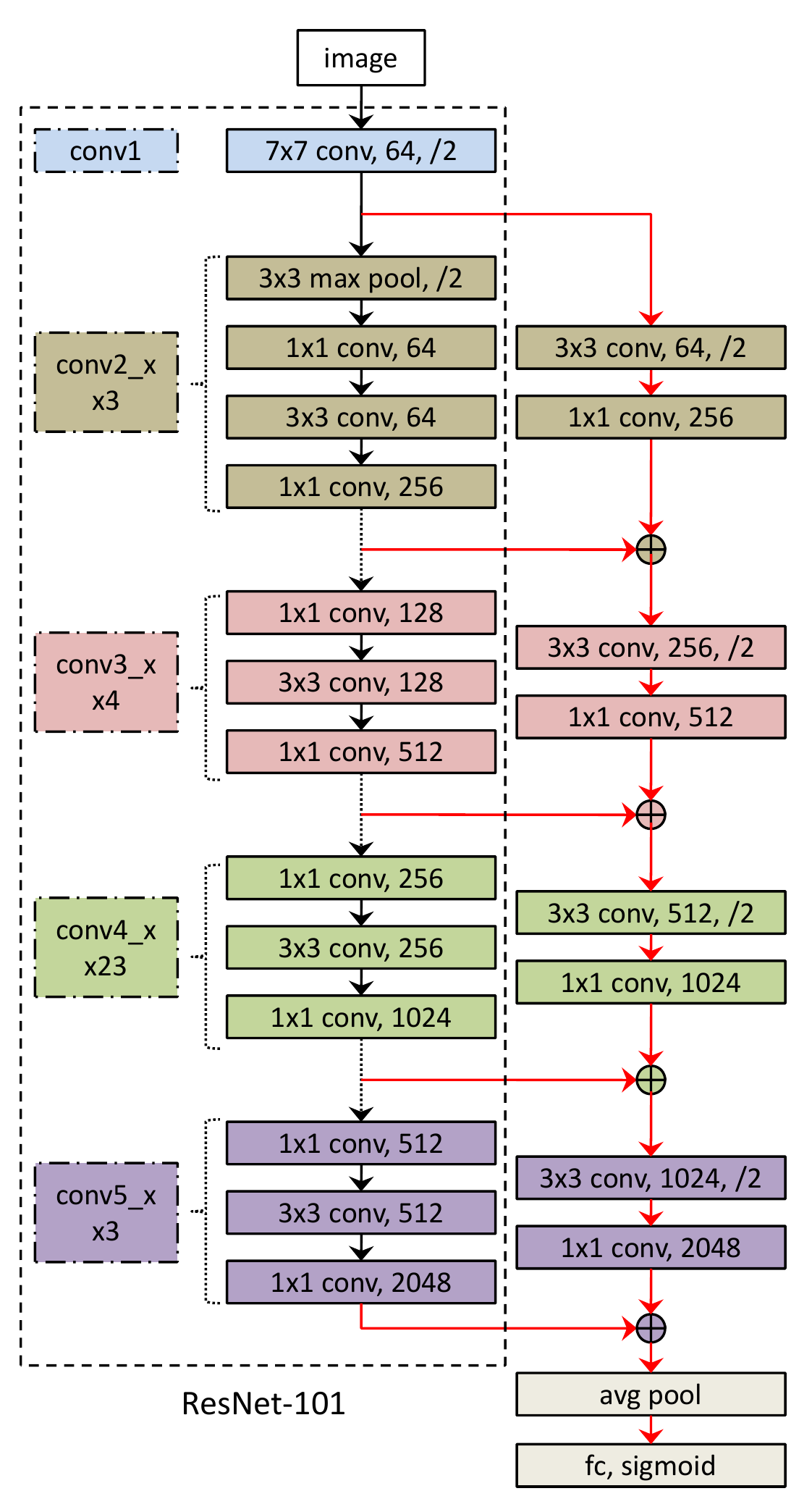}
\caption{The architecture of our multi-scale CNN subnetwork. Here, ResNet-101 is used as the basic CNN model, and four feature fusion blocks are included for multi-scale feature learning.}\label{fig4}
\end{figure}

In this paper, we take ResNet-101 \cite{he2016cvpr} as the main CNN branch, and select the final layers of five scales (\textit{i.e.}, conv1, conv2\_3, conv3\_4, conv4\_23 and conv5\_3, totally 5 final convolutional layers) for multi-scale fusion. In particular, the feature maps at the five scales have the size of $112\times112$, $56\times56$, $28\times28$, $14\times14$ and $7\times7$, along with $64$, $256$, $512$, $1024$ and $2048$ channels, respectively. The architecture of our multi-scale CNN subnetwork is shown in Fig.~\ref{fig4}.

\subsection{Multi-Layer Perception Model for Textual Feature Learning}

We further investigate how to learn textual features from noisy tags provided by users for social media websites. The noisy tags of image $\textbf{I}$ are represented as a binary vector $\bm{t}=\left(t_{1}, t_{2}, \cdots, t_{T}\right)$, where $T$ is the volume of tags, and $t_{j}=1$ if image $\textbf{I}$ is annotated with tag $j$. Since the binary vector is sparse and noisy, we encode the raw vector into a dense textual feature vector $\bm{f}_t$ using a multi-layer perception model, which consists of two hidden layers (each with 2,048 units), as shown in Fig.~\ref{fig2}. Note that only a simple neural network model is used for textual feature learning. Our consideration is that the noisy tags have a high-level semantic information and a complicated model would degrade the performance of textual feature learning. This observation has also been reported in \cite{Dhingra2016Tweet2Vec,Ling2015Two,Bai2014A} in the filed of natural language processing.

In this paper, the visual feature vector $\bm{f}_v$ and the textual feature vector $\bm{f}_t$ are forced to have the same dimension, which enables them to play the same important role in feature learning. By taking a multi-modal feature learning strategy, we concatenate the visual and textual feature vectors $\bm{f}_v$ and $\bm{f}_t$ into a joint feature vector $\bm{f}=[\bm{f}_v,\bm{f}_t]$ for the subsequent multi-class classification and label quantity prediction.

\subsection{Multi-Class Classification and Label Quantity Prediction}

\noindent \textbf{Multi-Class Classification.} Since each image can be annotated with one or more classes, we define a multi-class classifier for image annotation. Specifically, the joint visual and textual feature is connected to a fully connected layer for logit calculation, followed by a sigmoid function for probability calculation, as shown in Fig.~\ref{fig2}.\\
\textbf{Label Quantity Prediction.} We formulate the label quantity prediction problem as a regression problem, and adopt a multi-layer perception model as the regressor. As shown in Fig.~\ref{fig2}, the regressor consists of two hidden layers with $512$ and $256$ units, respectively. In this paper, to avoid the overfitting of the regressor, the dropout layers are applied in all hidden layers with the dropout rate 0.5.

\subsection{Model Training}

During the process of model training, we apply a multi-stage strategy and divide the architecture into several branches: visual features learning, textual features learning, multi-class classification, and label quantity prediction. Specifically, the original ResNet model is fine-tuned with a small learning rate. When the fine-tuning is finished, we will fix the parameters of ResNet and train the multi-scale blocks. In this paper, the training of the textual model is separated from the visual model, and thus the two models can be trained synchronously. After visual and textual feature learning, we fix the parameters of the visual and textual models, and train the multi-class classification and label quantity prediction models separately.

To provide further insights on model training, we define the loss functions for training the four branches as follows.\\
\textbf{Sigmoid Cross Entropy Loss.} For training the first three  branches, the features are first connected to a fully connected layer to compute the logits $\{z_{ij}\}$ for image $I_i$, and the sigmoid cross entropy loss is then applied for classification module training:
\begin{equation}
\mathcal{L}_{cls}=\sum_i\sum_{j}y_{ij}\times{\left(-\log\left(p_{ij}\right)\right)}+(1-y_{ij})\times{\left(-\log\left(1-p_{ij}\right)\right)}
\end{equation}
where $y_{ij}=1$ if image $I_i$ is annotated with class $j$, otherwise $y_{ij}=0$; and $p_{ij}=\sigma(z_{ij})$ with $\sigma(\cdot)$ as the sigmoid function. \\
\textbf{Mean Squared Error Loss.} For training the label quantity prediction model, the features are also connected to a fully connected layer with one output unit to compute the predicted label quantity $\hat{{m}_i}$ for image $I_i$. We then apply the following mean squared error loss for regression model training:
\begin{equation}
\mathcal{L}_{reg}=\sum_i\left(\hat{{m}_i}-{m_i}\right)^2
\end{equation}
where ${m_i}$ is the ground-truth label quantity of image $I_i$.

In this paper, the four branches of our model are not trained in a single process. Our main consideration is that our model has a over-complicated architecture and thus the overfitting issue still exists even if a large dataset is provided for model training. In the future work, we will make efforts to develop a robust algorithm for train our model end-to-end.

\subsection{Test Process}

At the test time, we first extract the joint visual and textual feature vector from each test image, and then execute multi-class classification and label quantity prediction synchronously. When the predicted probabilities of labels $\bm{p}$ and the predicted label quantity $\hat{m}$ have been obtained for the test image, we select the top $\hat{m}$ candidates as our final annotations.

\begin{figure*}[t]
\centering
\includegraphics[width=0.92\textwidth]{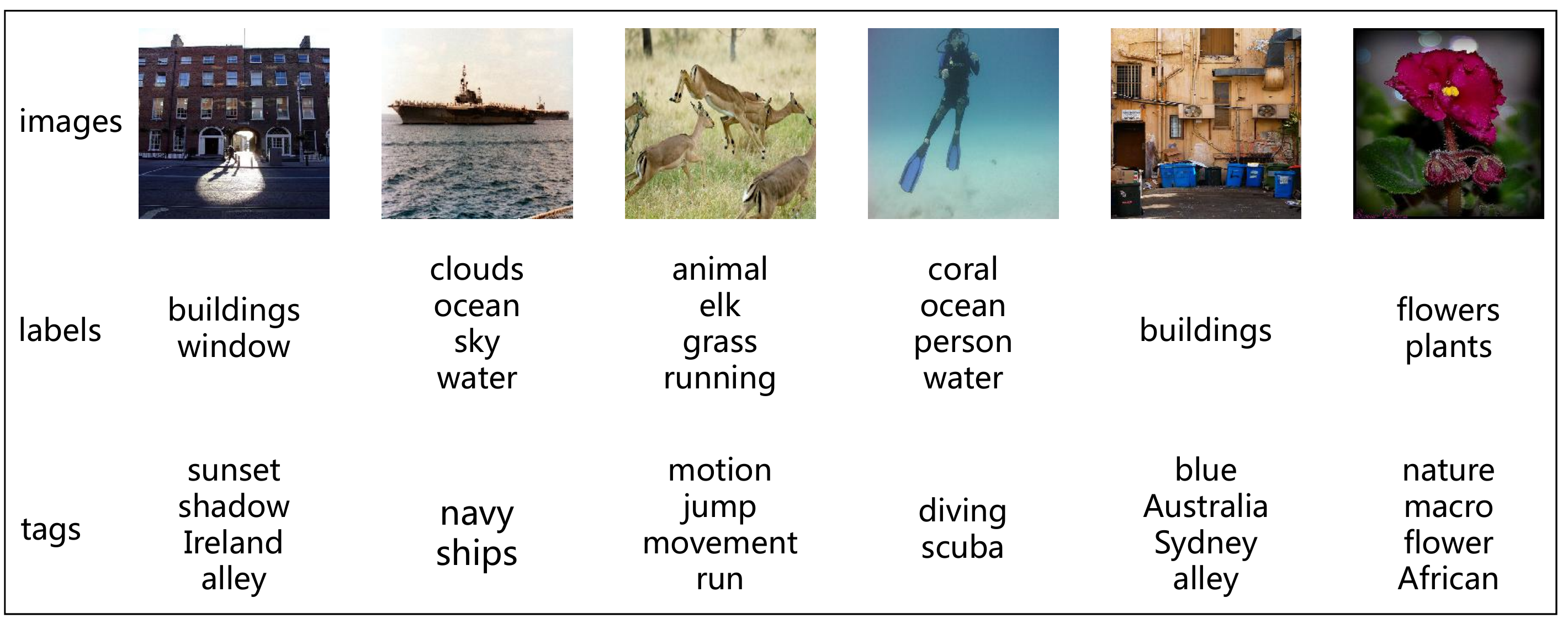}
\vspace{-0.05in}
\caption{Examples of the NUS-WIDE bechmark dataset. Images (first row) are followed with class labels (second row) and noisy tags (third row).}\label{fig5}
\vspace{-0.05in}
\end{figure*}

\section{Experiments}

\subsection{Datasets and Settings}

\subsubsection{Datasets}

\begin{figure}[t]
\centering
\includegraphics[width=0.48\textwidth]{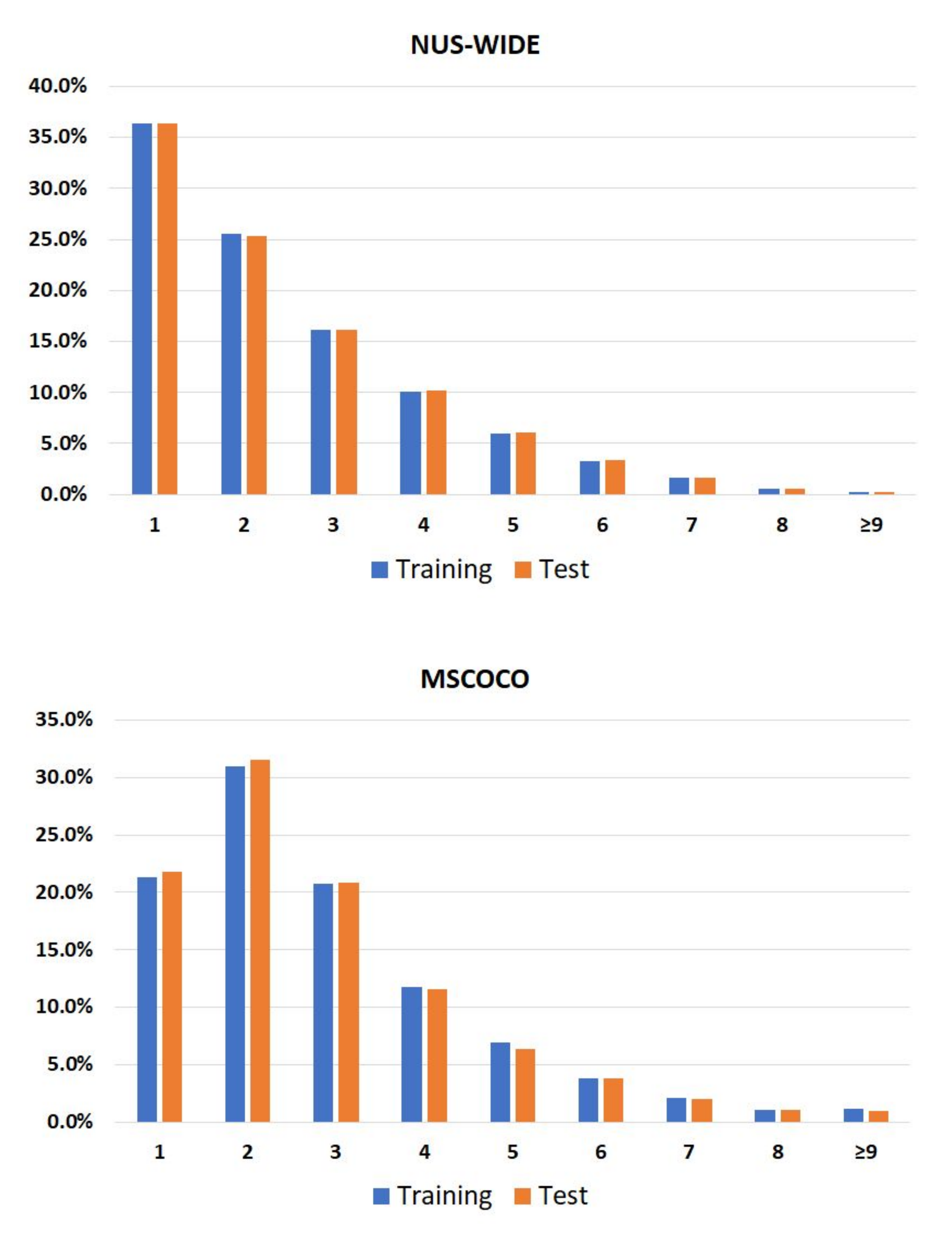}
\vspace{-0.05in}
\caption{The distribution of label quantity for NUS-WIDE and MSCOCO.}
\label{fig_lqp}
\vspace{-0.05in}
\end{figure}

The two most widely used benchmark datasets for large-scale image annotation are selected for performance evaluation. The first dataset is NUS-WIDE \cite{chua2009civr}, which consists of 269,648 images and 81 class labels from Flickr image metadata. The number of Flickr images varies in different studies, since some Flickr links are going invalid. For fair comparison, we also remove images without any social tag. As a result, we obtain 94,570 training images and 55,335 test images. Some examples of this dataset are shown in Fig.~\ref{fig5}. Over 20\% images from both training and test data have more than 3 labels, {while more than $35\%$ images have only one single label}. The maximum number of label quantity in the training and test data is 14 and 12, separately. The second dataset is MSCOCO \cite{vinyals2017pami}, which consists of 87,188 images and 80 class labels from the 2015 MSCOCO image captioning challenge. The training/test split of the MSCOCO dataset is 56,414/30,774. More than 25\% images from both training and test data have over 3 labels, {while over $20\%$ images have only one single label}. The maximum number of label quantity in the training and test data is 16 and 15, separately. {The visualized distribution of label quantity on the NUS-WIDE and MSCOCO datasets is shown in Fig. \ref{fig_lqp}.} In summary, there exists a distinct variety of label quantities in the two datasets, and thus it is necessary to predict label quantity for accurate and complete annotation. In addition, there are originally 5,018 and 170,339 noisy user tags released with NUS-WIDE and MSCOCO, respectively. We only keep 1,000 most frequent tags for each dataset as side information.

\subsubsection{Evaluation Metrics}

The per-class and per-image metrics including precision and recall have been widely used in related works \cite{gong2013arxiv,wang2016cvpr,jin2016icpr,johnson2015iccv,hu2016cvpr,liu2016arxiv}. In this paper, the per-class precision (C-P) and per-class recall (C-R) are obtained by computing the mean precision and recall over all the classes, while the overall per-image precision (I-P) and overall per-image recall (I-R) are computed by averaging over all the test images. Moreover, the per-class F1-score (C-F1) and overall per-image F1-score (I-F1) are used for comprehensive performance evaluation by combining precision and recall with the harmonic mean. The six metrics are defined as follows:
\begin{equation}
\begin{aligned}
&\textrm{C-P}=\frac{1}{C}\sum_{j=1}^{C}\frac{\mathrm{NI}_j^{c}}{\mathrm{NI}_j^{p}},\quad\textrm{I-P}=\frac{\sum_{i=1}^{N}\mathrm{NL}_i^{c}}{\sum_{i=1}^{N}\mathrm{NL}_i^{p}}\\
&\textrm{C-R}=\frac{1}{C}\sum_{j=1}^{C}\frac{\mathrm{NI}_j^{c}}{\mathrm{NI}_j^{g}},\quad\textrm{I-R}=\frac{\sum_{i=1}^{N}\mathrm{NL}_i^{c}}{\sum_{i=1}^{N}\mathrm{NL}_i^{g}}\\
&\textrm{C-F1}=\frac{2\cdot\textrm{C-P}\cdot\textrm{C-R}}{\textrm{C-P}+\textrm{C-R}},\quad\textrm{I-F1}=\frac{2\cdot\textrm{I-P}\cdot\textrm{I-R}}{\textrm{I-P}+\textrm{I-R}}
\end{aligned}
\end{equation}
where $C$ is the number of classes, $N$ is the number of test images, $\mathrm{NI}_j^{c}$ is the number of images correctly labelled as class $j$, $\mathrm{NI}_j^{g}$ is the number of images that have a ground-truth label of class $j$, $\mathrm{NI}_j^{p}$ is the number of images predicted as class $j$, $\mathrm{NL}_i^{c}$ is the number of correctly annotated labels for image $i$, $\mathrm{NL}_i^{g}$ is the number of ground-truth labels for image $i$, and $\mathrm{NL}_i^{p}$ is the number of predicted labels for image $i$.

According to \cite{gong2013arxiv}, the above per-class metrics are biased toward the infrequent classes, while the above per-image metrics are biased toward the frequent classes. Similar observations have also been presented in \cite{Lu2017Learning}. As a remedy, following the idea of \cite{Lu2017Learning}, we define a new metric called H-F1 with the harmonic mean of C-F1 and I-F1:
\begin{equation}
\quad\textrm{H-F1}=\frac{2\cdot\textrm{C-F1}\cdot\textrm{I-F1}}{\textrm{C-F1}+\textrm{I-F1}}
\end{equation}
Since H-F1 takes both per-class and per-image F1-scores into account, it enables us to make easy comparison between different methods for large-scale image annotation.

\begin{table*}[t]
\tabcolsep0.15cm
\centering
\caption{Effectiveness evaluation for the main components of our model on the NUS-WIDE dataset. } \label{tabNUS-WIDE1}
\vspace{-0.05in}
\begin{tabular}{l|c|c|ccc|ccc|c}
\hline
Models   & Multi-Modal & Quantity Prediction & C-P (\%)  & C-R (\%) & C-F1 (\%)  & I-P (\%)  & I-R (\%)  & I-F1 (\%) & H-F1 (\%)\\
\hline
Upper bound (Ours)  & yes & yes & 81.46 & 75.83 & 78.55 & 85.87 & 85.87 & 85.87 & 82.05 \\
\hline
MS-CNN+Tags+LQP  & yes & yes & \bf80.27 & \bf60.95 & \bf69.29 & \bf84.55 & \bf76.80 & \bf80.49 & \bf74.47 \\
MS-CNN+LQP & no & yes & 65.00 & 52.79 & 58.26 & 77.10 & 73.77 & 75.40 & 65.73\\
MS-CNN+Tags & yes & no & 57.15 & 64.39 & 60.55 & 62.85 & 74.03 & 67.98 & 64.05\\
MS-CNN   & no & no & 50.13 & 57.08 & 53.38 & 60.87 & 71.48 & 65.75 & 58.92\\
CNN  & no & no & 45.86 & 57.34 & 50.96 & 59.88 & 70.54 & 64.77 & 57.04\\
\hline
\end{tabular}
\vspace{-0.05in}
\end{table*}

\begin{table*}[t]
\tabcolsep0.15cm
\centering
\caption{Effectiveness evaluation for the main components of our model on the MSCOCO dataset. }
\label{tabMSCOCO1}
\vspace{-0.05in}
\begin{tabular}{l|c|c|ccc|ccc|c}
\hline
Models  & Multi-Modal & Quantity Prediction & C-P (\%)  & C-R (\%) & C-F1 (\%)  & I-P (\%)  & I-R (\%)  & I-F1 (\%) & H-F1 (\%) \\
\hline
Upper bound (Ours)  & yes & yes & 75.43 & 71.23 & 73.27 & 77.15 & 77.15 & 77.15 & 75.16 \\
\hline
MS-CNN+Tags+LQP & yes & yes & \bf74.88 & \bf64.96 & \bf69.57 & \bf76.34 & \bf70.25 & \bf73.17 & \bf71.32 \\
MS-CNN+LQP  & no & yes & 67.48 & 60.93 & 64.04 & 70.22 & 67.93 & 69.06 & 66.46 \\
MS-CNN+Tags & yes & no & 63.69 & 63.87 & 63.78 & 64.47 & 68.77 & 66.55 & 65.14 \\
MS-CNN  & no & no & 59.23 & 59.89 & 59.56 & 62.08 & 66.11 &	64.03 &	61.71 \\
CNN     & no & no & 57.16 & 57.24 & 57.20 & 60.23 & 64.25 & 62.17 & 59.58 \\
\hline
\end{tabular}
\vspace{-0.05in}
\end{table*}

\subsubsection{Settings}

In this paper, the basic CNN module makes use of ResNet-101 \cite{he2016cvpr} which is pretrained on the ImageNet 2012 classification challenge dataset \cite{Russakovsky2014ImageNet}. Our experiments are all conducted on TensorFlow. The input images are resized to 224$\times$224 pixels. For training the basic CNN model, the multi-scale CNN model, and the multi-class classifier, the learning rate is set to 0.001 for NUS-WIDE and 0.002 for MSCOCO, respectively. {The learning rate is multiplied by 0.1 after every 80,000 iterations, up to 160,000 iterations with early stopping.} For training the textual feature learning model and the label quantity prediction model, the learning rate is set to 0.1 for NUS-WIDE and 0.0001 for MSCOCO, respectively. These models are trained using a Momentum with the momentum rate of 0.9. The decay rate of batch normalization and weight decay is set to 0.9997.

\subsubsection{Compared Methods}

We conduct two groups of experiments for evaluation and choose competitors to compare accordingly: (1) We first make comparison among several variants of our complete model shown in Fig.~\ref{fig2} by removing one or more components from our model, so that the effectiveness of each component can be evaluated properly. (2) To compare with a wider range of image annotation methods, we also compare with the published results on the two benchmark datasets. These include the state-of-the-art deep learning methods for large-scale image annotation.

\vspace{-0.15cm}
\subsection{Effectiveness Evaluation for Model Components}

We conduct the first group of experiments to evaluate the effectiveness of the main components of our model for large-scale image annotation. Five closely related models are included in component effectiveness evaluation: (1) CNN denotes the original ResNet-101 model; (2) MS-CNN denotes the multi-scale ResNet model shown in Fig.~\ref{fig3}; (3) MS-CNN+Tags denotes the multi-modal multi-scale ResNet model that learns both visual and textual features for image annotation; (4) MS-CNN+LQP denotes the multi-scale ResNet model that performs both multi-class classification and label quantity prediction (LQP) for image annotation; (5) MS-CNN+Tags+LQP denotes our complete model shown in Fig.~\ref{fig2}. Note that the five models can be recognized by whether they are multi-scale, multi-modal, or making LQP. This enables us to evaluate the effectiveness of each component of our model.

The results on the two benchmark datasets are shown in Tables I and II, respectively. Here, we also show the upper bounds of our complete model (i.e. MS-CNN+Tags+LQP) obtained by directly using the ground-truth label quantities to refine the predicted annotations (without LQP). We can make the following observations: (1) Since label quantity prediction is explicitly addressed in our model (unlike RNN), it indeed leads to significant improvements according to the H-F1 score (10.42 percent for NUS-WIDE, and 6.18 percent for MSCOCO), when MS-CNN+Tags is used for feature learning. The improvements achieved by label quantity prediction become smaller when only MS-CNN is used for feature learning. This is because that the quality of label quantity prediction would degrade when the social tags are not used as textual features. (2) The social tags are crucial not only for label quantity prediction but also for the final image annotation. Specifically, the textual features extracted from social tags yield significant gains according to the H-F1 score (8.74 percent for NUS-WIDE, and 4.86 percent for MSCOCO), when both MS-CNN and LQP are adopted for image annotation. This is also supported by the gains achieved by MS-CNN+Tags over MS-CNN. (3) The effectiveness of MS-CNN is verified by the comparison MS-CNN vs. CNN. Admittedly  only small gains ($>1\%$ in terms of H-F1) have been obtained by MS-CNN. However, this is still impressive since ResNet-101 is a very powerful CNN model. In summary, we have evaluated the effectiveness of all the components of our complete model (shown in Fig.~\ref{fig2}).

\begin{table*}[t]
\tabcolsep0.15cm
\centering
\caption{Comparison to the state-of-the-art on the NUS-WIDE dataset. }
\label{tabNUS-WIDE2}
\vspace{-0.05in}
\begin{tabular}{l|c|c|ccc|ccc|c}
\hline
Methods   & Multi-Modal & Quantity Prediction & C-P (\%)  & C-R (\%) & C-F1 (\%)  & I-P (\%)  & I-R (\%)  & I-F1 (\%) & H-F1 (\%)\\
\hline
Upper bound (Ours)  & yes & yes & 81.46 & 75.83 & 78.55 & 85.87 & 85.87 & 85.87 & 82.05 \\
\hline
Ours  & yes & yes  & \bf80.27 & 60.95 & \bf69.29 & \bf84.55 & 76.80 & \bf80.49 & \bf74.47 \\
SR-CNN-RNN \cite{liu2016arxiv}  & yes & yes  & 71.73 & \bf61.73 & 66.36 & 77.41 & 76.88 & 77.15 & 71.35 \\
SINN \cite{hu2016cvpr}   & yes & yes  & 58.30 & 60.30 & 59.44 & 57.05 & \bf79.12 & 66.30 & 62.68 \\
TagNeighboor \cite{johnson2015iccv} & yes & no & 54.74 & 57.30 & 55.99 & 53.46 & 75.10 & 62.46 & 59.05 \\
RIA   \cite{jin2016icpr}  & no &  yes & 52.92 & 43.62 & 47.82 & 68.98 & 66.75 & 67.85  & 56.10 \\
CNN-RNN  \cite{wang2016cvpr}  & no & no   & 40.50 & 30.40 & 34.70 & 49.90 & 61.70 & 55.20 & 42.61 \\
CNN+WARP \cite{gong2013arxiv}   & no & no  & 31.65 & 35.60 & 33.51 & 48.59 & 60.49 & 53.89  & 41.32 \\
CNN+softmax \cite{gong2013arxiv}  & no & no & 31.68 & 31.22 & 31.45 & 47.82 & 59.52 & 53.03 & 39.48 \\
CNN+logistic \cite{hu2016cvpr}     & no & no & 45.60 & 45.03 & 45.31 & 51.32 & 70.77 & 59.50  & 51.44 \\
\hline
\end{tabular}
\vspace{-0.05in}
\end{table*}

\begin{table*}[t]
\tabcolsep0.15cm
\centering
\caption{Comparison to the state-of-the-art on the MSCOCO dataset. }
\label{tabMSCOCO2}
\vspace{-0.05in}
\begin{tabular}{l|c|c|ccc|ccc|c}
\hline
Methods    & Multi-Modal & Quantity Prediction   & C-P (\%)  & C-R (\%) & C-F1 (\%)  & I-P (\%)  & I-R (\%)  & I-F1 (\%) & H-F1 (\%) \\
\hline
Upper bound (Ours)  & yes & yes & 75.43 & 71.23 & 73.27 & 77.15 & 77.15 & 77.15 & 75.16  \\
\hline
Ours & yes & yes & \bf74.88 & \bf64.96 & \bf69.57 & 76.34 & 70.25 & 73.17 & \bf71.32 \\
SR-CNN-RNN \cite{liu2016arxiv}   & yes & yes & 71.38 & 63.13 & 67.00 & \bf77.41 & \bf73.05 & \bf75.16 & 70.84 \\
RIA \cite{jin2016icpr}     & no & yes   & 64.32 & 54.07 & 58.75 & 74.20 & 64.57 & 69.05 & 63.48 \\
CNN-RNN  \cite{wang2016cvpr}     & no &  no & 66.00 & 55.60 & 60.40 & 69.20 & 66.40 & 67.80 & 63.89 \\
CNN+WARP \cite{wang2016cvpr}     & no &  no & 52.50 & 59.30 & 55.70 & 61.40 & 59.80 & 60.70 & 58.09 \\
CNN+softmax \cite{wang2016cvpr}   & no & no & 57.00 & 59.00 & 58.00 & 62.10 & 60.20 & 61.10 & 59.51 \\
CNN+logistic \cite{wang2016cvpr}   & no & no & 59.30 & 58.60 & 58.90 & 61.70 & 65.00 & 63.30 & 61.02 \\
\hline
\end{tabular}
\vspace{-0.05in}
\end{table*}

\vspace{-0.2cm}
\subsection{Comparison to the State-of-the-Art Methods}

In this group of experiments, we compare our method with the state-of-the-art deep learning methods for large-scale image annotation. The following competitors are included: (1) CNN+Logistic \cite{hu2016cvpr}: This model fits a logistic regression classifier for each class label. (2) CNN+Softmax \cite{gong2013arxiv}: It is a CNN model that uses softmax as classifier and the cross entropy as the loss function. (3) CNN+WARP \cite{gong2013arxiv}: It uses the same CNN model as above, but a weighted approximate ranking loss function is adopted for training to promote the prec@K metric. (4) CNN-RNN \cite{wang2016cvpr}: It is a CNN-RNN model that uses output fusion to merge CNN output features and RNN outputs. (5) RIA \cite{jin2016icpr}: In this CNN-RNN model, the CNN output features are used to set the hidden state of Long short-term memory (LSTM) \cite{HS97}. (6) TagNeighbour \cite{johnson2015iccv}: It uses a non-parametric approach to find image neighbours according to metadata, and then aggregates image features for classification. (7) SINN \cite{hu2016cvpr}: It uses different concept layers of tags, groups, and labels to model the semantic correlation between concepts of different abstraction levels, and a bidirectional RNN-like algorithm is developed to integrate information for prediction. (8) SR-CNN-RNN \cite{liu2016arxiv}: This CNN-RNN model uses a semantically regularised embedding layer as the interface between the CNN and RNN.

The results on the two benchmark datasets are shown in Tables III and IV, respectively. Here, we also show the upper bounds of our model obtained by directly using the ground-truth label quantities to refine the predicted annotations (without LQP). It can be clearly seen that our model outperforms the state-of-the-art deep learning methods according to the H-F1 score. This provides further evidence that our multi-modal multi-scale CNN model along with label quantity prediction is indeed effective in large-scale image annotation. Moreover, it is also noted that our model with explicit label quantity prediction yields better results than the CNN-RNN models with implicit label quantity prediction (i.e. SR-CNN-RNN, RIA, and SINN). This shows that RNN is not the unique model suitable for label quantity prediction. In particular, when it is done explicitly like our model, we can pay more attention to the CNN model itself for deep feature learning. Considering that RNN needs prior knowledge on class labels, our model is expected to have a wider use in real-word applications. In addition, the annotation methods that adopt multi-modal feature learning or label quantity prediction are generally shown to outperform the methods without considering any of the two components for image annotation.

\begin{table}[t]
\tabcolsep0.5cm
\centering
\caption{Comparison of different multi-scale CNN models on the two datasets. H-F1 (\%) is used.}
\label{tabNUSCOCOMS}
\vspace{-0.05in}
\scalebox{0.88}{\begin{tabular}{l|c|c|c}
\hline
Models   &  Dimension  & NUS-WIDE  & MSCOCO \\
\hline
MS-CNN    & 2,048  & \textbf{58.92} & \textbf{61.71} \\
MS-CNN-AvgPool  & 3,904 & 57.93 & 60.73 \\
MS-CNN-MaxPool & 2,048 & 58.04 & 60.67 \\
CNN   & 2,048  & 57.04 & 59.58 \\
\hline
\end{tabular}}
\vspace{-0.05in}
\end{table}

\begin{table}[t]
\centering
\caption{Performances of MS-CNN models fusing different scales of features on the NUS-WIDE and MSCOCO dataset. H-F1 (\%) is used.}
\label{tabAblation}
\vspace{-0.05in}
\scalebox{0.9}{\begin{tabular}{c|c|c|c|c}
\hline
Models & Fusion & Fused Layers & NUS-WIDE & MSCOCO\\
\hline
MS-CNN  & sum & 1,2,3,4,5 & \textbf{58.92} & \textbf{61.71}\\
MS-CNN  & sum & 2,3,4,5 & 58.69 & 61.43\\
MS-CNN  & sum & 3,4,5 & 58.33 & 60.96\\
MS-CNN  & sum & 4,5 & 57.89 & 60.74\\
\hline
MS-CNN-Avgpool & concat & 1,2,3,4,5 & 57.93 & 60.73 \\
MS-CNN-Avgpool & concat & 2,3,4,5 & 57.91 & 60.72 \\
MS-CNN-Avgpool & concat & 3,4,5 & 57.88 & 60.71 \\
MS-CNN-Avgpool & concat & 4,5 & 57.86 & 60.68 \\
\hline
MS-CNN-Maxpool & sum & 1,2,3,4,5 & 58.04 & 60.67 \\
MS-CNN-Maxpool & sum & 2,3,4,5 & 57.76 & 60.41\\
MS-CNN-Maxpool & sum & 3,4,5 & 57.53 & 60.04 \\
MS-CNN-Maxpool & sum & 4,5 & 57.31 & 59.83 \\
\hline
CNN	    & - & 5 & 57.04 & 59.58\\
\hline
\end{tabular}}
\vspace{-0.05in}
\end{table}

\begin{table}[t]
\centering
\caption{Performance of our multi-modal model using different textual features on the two datasets. H-F1 (\%) is used.}
\label{tabNUS-COCOw2v}
\begin{tabular}{c|c|c}
\hline
Model & NUS-WIDE  & MSCOCO \\
\hline
MS-CNN+MLP      &  \bf 64.05 & \bf 65.14\\
\hline
MS-CNN+Word2vec &  62.17  & 63.17 \\
\hline
\end{tabular}
\vspace{-0.05in}
\end{table}

\begin{table}[t]
\tabcolsep0.55cm
\centering
\caption{Results of label quantity prediction (LQP) with different groups of features on the NUS-WIDE dataset.}
\label{tabNUS-WIDELQP}
\vspace{-0.05in}
\begin{tabular}{l|cc}
\hline
Features for LQP   & Accuracy (\%)  & MSE \\
\hline
MS-CNN      & 45.92 & 0.673 \\
MS-CNN+Tags  & \bf48.26 & \bf0.437 \\
\hline
\end{tabular}
\vspace{-0.05in}
\end{table}

\begin{table}[t]
\tabcolsep0.55cm
\centering
\caption{Results of label quantity prediction with different groups of features on the MSCOCO dataset. }
\label{tabMSCOCOLQP}
\vspace{-0.05in}
\begin{tabular}{l|cc}
\hline
Features for LQP      & Accuracy (\%)  & MSE \\
\hline
MS-CNN       & 46.27 & 0.773 \\
MS-CNN+Tags  & \bf47.96 & \bf0.564 \\
\hline
\end{tabular}
\vspace{-0.05in}
\end{table}

\subsection{Further Evaluation}

\subsubsection{Alternative Multi-Scale CNN Models}

We also make comparison to two variant versions of our multi-scale CNN model (denoted as MS-CNN): 1) MS-CNN-AvgPool that directly combines the final layers of five scales (i.e. conv1, conv2\_3, conv3\_4, conv4\_23 and conv5\_3) using average pooling, resulting in a 3,904-dimensional feature vector; 2) MS-CNN-MaxPool that only replaces the $3\times 3$ conv in Fig. 3 by max pooling, without changing any of the other parts of our MS-CNN. The results on the two benchmark datasets are shown in Table \ref{tabNUSCOCOMS}. It is clearly shown that all the multi-scale CNN models outperforms the baseline CNN model (i.e. ResNet-101), due to multi-scale feature learning. {Moreover, the multi-scale fusion method used in our MS-CNN is shown to yield about \%1 gains over those used in MS-CNN-AvgPool and MS-CNN-MaxPool. In particular, as compared to MS-CNN-AvgPool, our MS-CNN does not increase the feature dimension and thus is more practicable.}

In addition, we provide the ablative results using different fusion methods in Table \ref{tabAblation}. Here, ``sum'' represents element-wise feature map sum, and ``concat'' means that we directly concatenate different levels of features after down sampling. With lower-level features fused, the performance (H-F1) of the ``sum'' fusion increases linearly, while the performance of the ``concat'' fusion accelerates insignificantly. This shows that the ``sum'' fusion benefits more from low-level features while maintaining the same feature dimension. The reason is that simply down sampling and concatenating low-level feature maps just throw out many pixels and make little use of low-level information such as texture and shape.

\subsubsection{Alternative Textual Features}

We explore different types of sub-networks for textual features extraction from noisy tags. Word2vec \cite{witten2016data} is one of the most popular methods for textual feature extraction. Since the 1,000 most frequent tags are out of order and cannot form meaningful sentences, it is hard to directly generate word2vec features from these tags. Instead, we compare MLP textual features with GloVe \cite{pennington2014glove} word vectors, which are pre-trained on the Wikipedia 2014 + Gigaword 5 dataset and then fine-tuned on the two datasets. Specifically, we encode each tag into an embedding vector, and accumulate those vectors for each image into the final word2vec features. As shown in Table \ref{tabNUS-COCOw2v}, our MLP features outperform the word2vec features by about 2 percent on the two datasets. The possible reason is that the word2vec features are more sensitive to the correctness of tag words. The attention mechanism can help to extract more stable word2vec features, which is not the focus of this paper.

\subsubsection{Results of Label Quantity Prediction}

We have evaluated the effectiveness of LQP in the above experiments, but have not shown the quality of LQP itself. In this paper, to measure the quality of LQP, two metrics are computed as follows: 1) Accuracy: the predicted label quantities are first quantized to their nearest integers, and then compared to the ground-truth label quantities to obtain the accuracy; 2) Mean Squared Error (MSE): the mean squared error is computed by directly comparing the predicted label quantities to the ground-truth ones. The results of LQP on the two benchmark datasets are shown in Tables \ref{tabNUS-WIDELQP} and \ref{tabMSCOCOLQP}, respectively. It can be seen that: (1) More than 45\% label quantities are correctly predicted in all cases. (2) The textual features extracted from social tags yield significant gains when MSE is used as the measure. {In addition, our LQP method is also shown to be able to handle the extreme case when the quantity number is large or small. For images which have only one single label, the predicted label quantity is $1.20\pm0.52$ on NUS-WIDE and $1.16\pm0.47$ on MSCOCO, which means that our LQP method can handle the case when the label quantity reaches the minimum number. For the few images which have more than 8 labels, the predicted label quantity on NUS-WIDE and MSCOCO is $6.04\pm1.34$ and $6.32\pm1.12$, respectively. Note that these images with large quantity number are rare in both training and test sets ($<0.5\%$ for NUS-WIDE and $<1.5\%$ for MSCOCO), and thus it is difficult for any model to precisely estimate the label quantity of outliers. Although the predicted number is only a conservative estimation of ground-truth label quantity, our LQP metod can still capture the important information that there exist multiple objects in these images.}

\begin{table}[t]
\centering
\caption{Performance of different label quantity determination methods on NUS-WIDE and MSCOCO. ``LQP'' denotes our label quantity prediction method, and ``Threshold-$p$'' means taking the threshold of classification probability as $p$ for all images. H-F1 (\%) is used.}
\label{tabTOP}
\vspace{-0.05in}
\begin{tabular}{c|c|c}
\hline
Models & NUS-WIDE & MSCOCO\\
\hline
MS-CNN+LQP  & \textbf{65.73} & \textbf{66.46}\\
\hline
MS-CNN & 58.92 & 61.71 \\
\hline
MS-CNN+Threshold 0.7 & 55.17 & 56.53 \\
MS-CNN+Threshold 0.5 & 60.46 & 61.29 \\
MS-CNN+Threshold 0.3 & 63.67 & 64.32 \\
MS-CNN+Threshold 0.1 & 58.92 & 60.14 \\
\hline
\end{tabular}
\vspace{-0.1in}
\end{table}

\begin{table}[t]
\centering
\caption{Comparison of different LQP methods on NUS-WIDE and MSCOCO. H-F1 (\%) is used.}
\label{tabBIN}
\vspace{-0.05in}
\begin{tabular}{c|c|c|c}
\hline
Models & LQP  & NUS-WIDE & MSCOCO\\
\hline
MS-CNN+LQP & Regression & \bf 65.73 & \bf 66.46 \\
MS-CNN+LQP & Classification & 62.67 & 64.28\\
MS-CNN     & No   & 58.92 & 61.71 \\
\hline
\end{tabular}
\vspace{-0.1in}
\end{table}

\begin{table}[t]
\centering
\caption{Performance of our model trained in the end-to-end manner on NUS-WIDE and MSCOCO. H-F1 \% is used.}
\label{tabEND2END}
\vspace{-0.05in}
\scalebox{0.9}{\begin{tabular}{c|c|c|c}
\hline
Models & End-to-End & NUS-WIDE & MSCOCO\\
\hline
MS-CNN+Tags+LQP & Yes & \bf 75.13 & \bf 72.03 \\
MS-CNN+Tags+LQP & No  & 74.47 & 71.32 \\
\hline
\end{tabular}}
\vspace{-0.1in}
\end{table}

\begin{figure*}[t]
\centering
\includegraphics[width=0.85\textwidth]{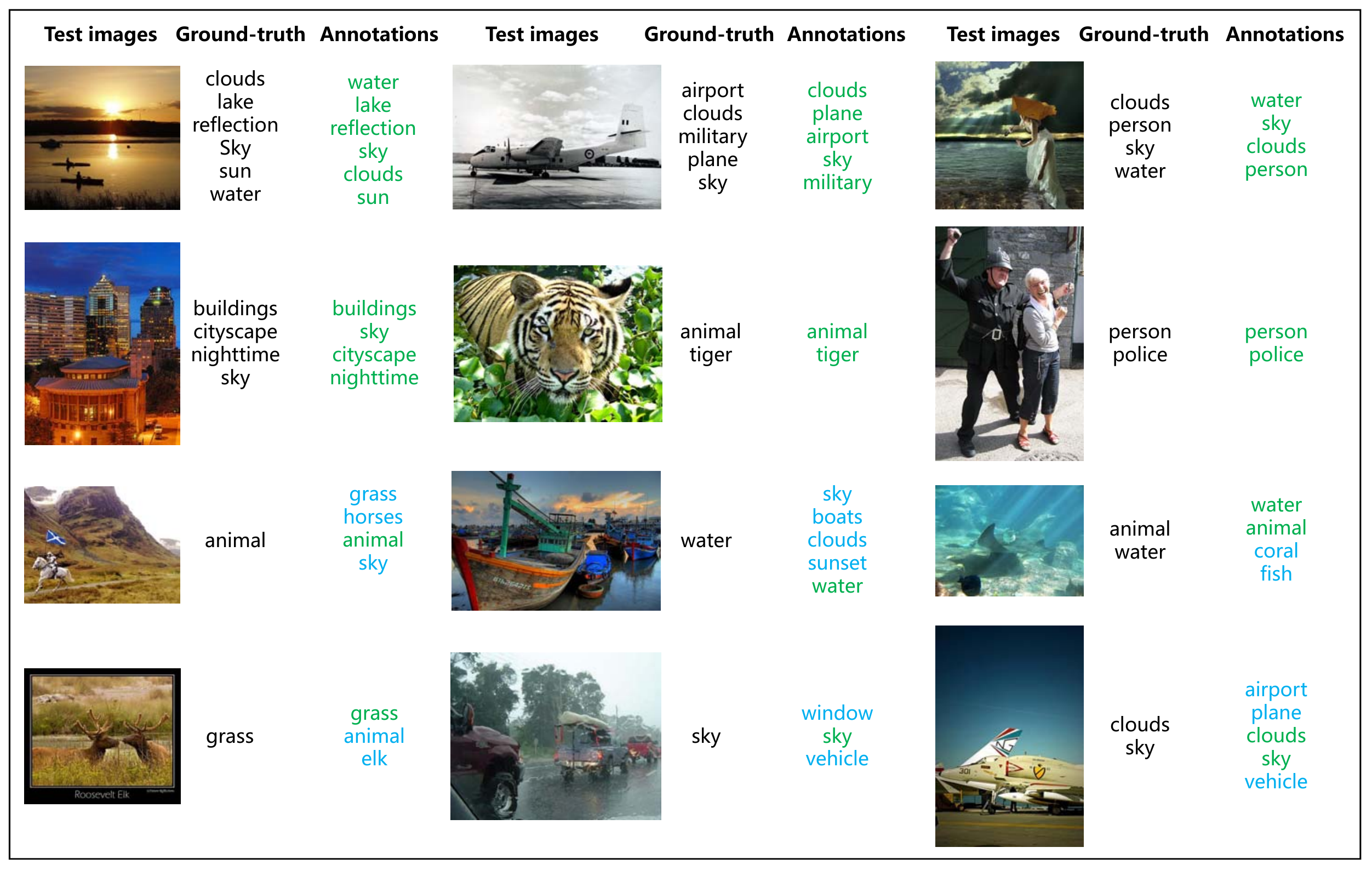}
\vspace{-0.1in}
\caption{Example results on NUS-WIDE obtained by our model. Annotations with green font are included in ground-truth class labels, while annotations with blue font are correctly tagged but not included in ground-truth class labels.}
\label{fig6}
\vspace{-0.1in}
\end{figure*}

We also explore other label quantity determination methods. As shown in Table \ref{tabTOP}, taking a proper threshold (e.g. $p=0.3$) of classification probability for determining the optimal label number of each image can help to yield better annotations, as compared to the conventional top-$k$ label selection method (i.e. ``MS-CNN''). However, it is hard to select the optimal threshold for label quantity determination. Therefore, our LQP method is clearly shown to outperform the threshold method.

Another candidate method for label quantity prediction is ``Classification'', which regards each number of label quantity as one category and applies softmax cross entropy for training. Note that our LQP method is different from this ``Classification'' method in that label quantity prediction is formulated as a regression problem using mean square error loss (thus denoted as ``Regression''). The experimental results in Table \ref{tabBIN} demonstrate that label quantity prediction consistently leads to performance improvements no matter which LQP method is adopted. Moreover, the ``Regression'' method outperforms ``Classification'' by over 2\%, which indicates that the ``Classification'' method is not a sound choice for label quantity prediction. The explanation is that the huge gap between the ground-truth quantity and predicted number can not be properly penalized by the classification loss. In contrast, the mean square error used in our LQP model can ensure that the prediction is not far away from the ground-truth quantity.

\subsubsection{End-to-End Training}

We can also train our whole model in an end-to-end manner. Specifically, the visual and textual representation learning module (MS-CNN+Tags) is pre-trained as initialization, and then we fine-tune all parameters except vanilla ResNet. The objective function is the sum of multi-label classification loss and label quantity regression loss. The comparative results in Table \ref{tabEND2END} show the effectiveness of end-to-end training, which means that both classification and label quantity prediction can be promoted from each other, and a better multi-modal feature representation is achieved by interaction between visual and textual branches.

\subsubsection{Qualitative Results of Image Annotation}

Fig.~\ref{fig6} shows several annotation examples on NUS-WIDE when our model is employed. Here, annotations with green font are included in ground-truth class labels, while annotations with blue font are correctly tagged but not included in ground-truth class labels (i.e. the ground-truth labels may be incomplete). It is observed that the images in the first two rows are all annotated exactly correctly, while this is not true for the images in the last two rows. However, by checking the extra annotations (with blue font) generated for the images in the last two rows, we find that they are all consistent with the content of the images.

\section{Conclusion}

We have proposed a novel multi-modal multi-scale deep learning model for large-scale image annotation. Compared to existing models, the main difference is the multi-scale feature learning architecture designed to extract and fuse features at different layers suitable for representing visual concepts of different levels of abstraction. Furthermore,  different from the RNN-based models with implicit label quantity prediction, a regressor is directly added to our model for explicit label quantity prediction. Extensive experiments are carried out to demonstrate that the proposed model outperforms the state-of-the-art methods. Moreover, each component of our model has also been shown to be effective in large-scale image annotation. A number of directions are worth further investigation. Firstly, the current multi-scale feature fusion architecture is based on the ResNet structure; it is desirable to investigate how other types of architecture can in integrated into our framework for multi-scale feature fusion.  Secondly, other types of side information (e.g. group labels) can be fused in our model. Finally, extracting and fusing multi-scale deep features are also important for other visual recognitions tasks which require multi-scale or spatial layout sensitive representation. Part of the ongoing work is thus to investigate how the proposed model can be applied to those tasks.

\section*{Acknowledgement}

This work was partially supported by National Natural Science Foundation of China (61573363), the Fundamental Research Funds for the Central Universities and the Research Funds of Renmin University of China (15XNLQ01), and the Outstanding Innovative Talents Cultivation Funded Programs 2016 of Renmin University of China.


\end{document}